\documentclass[11pt]{article}
\usepackage{enumitem}
\usepackage[final]{acl}
\usepackage{amssymb}
\usepackage{algorithm}
\usepackage{algorithmic}
\usepackage{amsmath}
\usepackage{placeins}
\usepackage{float}
\usepackage{booktabs}
\usepackage{array}
\usepackage{times}
\usepackage{latexsym}

\usepackage[T1]{fontenc}

\usepackage[utf8]{inputenc}

\usepackage{microtype}

\usepackage{inconsolata}

\usepackage{graphicx}

\title{ASLoRA: Adaptive Sharing Low-Rank Adaptation Across Layers}

\author{
 \textbf{Junyan Hu\textsuperscript{1}},
 \textbf{Xue Xiao\textsuperscript{2}},
 \textbf{Mengqi Zhang\textsuperscript{1}},
 \textbf{Yao Chen\textsuperscript{2}},
 \textbf{Zhaochun Ren\textsuperscript{3}},\\
 \textbf{Zhumin Chen\textsuperscript{1}},
 \textbf{Pengjie Ren\textsuperscript{1}\thanks{Corresponding author}}
\\
 \textsuperscript{1}Shandong University, Qingdao, China\\
 \textsuperscript{2}Inspur Cloud Information Technology Co.,Ltd\\
 \textsuperscript{3}Leiden University, Leiden, The Netherlands
\\
 \texttt{hujunyan@mail.sdu.edu.cn}\\
 \texttt{\{renpengjie,mengqi.zhang,chenzhumin\}@sdu.edu.cn}\\
 \texttt{xiaoxue@inspur.com, chenyao@inspur.com}
 }

\begin{document}
\maketitle
\begin{abstract}
As large language models (LLMs) grow in size, traditional full fine-tuning becomes increasingly impractical due to its high computational and storage costs. Although popular parameter-efficient fine-tuning methods, such as LoRA, have significantly reduced the number of tunable parameters, there is still room for further optimization. In this work, we propose ASLoRA, a cross-layer parameter-sharing strategy combining global sharing with partial adaptive sharing. Specifically, we share the low-rank matrix $A$ across all layers and adaptively merge matrix $B$ during training. This sharing mechanism not only mitigates overfitting effectively but also captures inter-layer dependencies, significantly enhancing the model's representational capability. We conduct extensive experiments on various NLP tasks, showing that ASLoRA outperforms LoRA while using less than 25\% of the parameters, highlighting its flexibility and superior parameter efficiency. Furthermore, in-depth analyses of the adaptive sharing strategy confirm its significant advantages in enhancing both model flexibility and task adaptability.
\end{abstract}

\section{Introduction}
The advent of large language models (LLMs) like GPT-3.5 Turbo~\cite{DBLP:journals/corr/abs-2303-08774}, Gemini~\cite{DBLP:journals/corr/abs-2312-11805}, and LLaMA3~\cite{DBLP:journals/corr/abs-2407-21783} marks a breakthrough in NLP. However, due to their massive parameters, fully fine-tuning these models for specific tasks is expensive, especially as model sizes grow~\cite{10.5555/3495724.3495883}. In response, parameter-efficient fine-tuning (PEFT), such as adapter~\cite{DBLP:conf/icml/HoulsbyGJMLGAG19, hu2022lora} and Prefix Tuning~\cite{li-liang-2021-prefix}, have gained popularity. These methods fine-tune only a small subset of parameters, reducing storage and computation demands significantly.

As a popular method of parameter-efficient fine-tuning (PEFT), LoRA~\cite{hu2022lora} introduces two low-rank matrices, $A$ and $B$, whose product represents the update to the weight matrix, i.e., $W_0+\Delta W=W_0+B A$. Given that the ranks of $A$ and $B$ are significantly smaller than the original model dimensions, this approach greatly reduces the number of tunable parameters. Moreover, LoRA directly adds the product of the low-rank matrices to the weight matrix, without introducing additional inference latency. Despite its excellent performance, LoRA still requires a substantial number of parameters.

\begin{figure}[t!]
    \centering
    \begin{minipage}{0.25\textwidth}
        \centering
        \includegraphics[width=\textwidth]{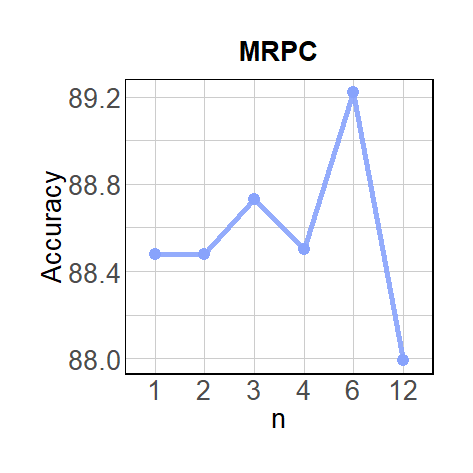}
    \end{minipage}%
    \begin{minipage}{0.25\textwidth}
        \centering
        \includegraphics[width=\textwidth]{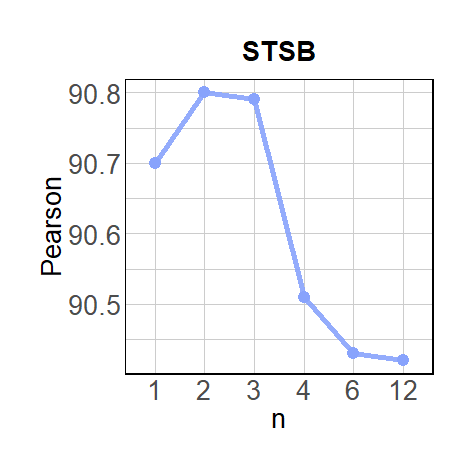}
    \end{minipage}
    \caption{The pre-experiment on the MRPC and STSB datasets. We let $A$ be shared in all layers and make adjacent $n$ layers share the same $B$, where $n=3$ means that every 3 adjacent layers share the same $B$.}
    \label{fig:pre}
\end{figure}
To address this issue, several studies have explored combining parameter sharing with LoRA. For instance, VeRA~\cite{DBLP:conf/iclr/KopiczkoBA24} shares randomly initialized matrices $A$ and $B$ across all layers and freezes their parameters while introducing trainable scaling vectors between them to reduce the number of tunable parameters. However, their weight-freezing strategy limits model expressiveness. Subsequently, Tied LoRA~\cite{renduchintala-etal-2024-tied} alleviates these issues by allowing trainable matrices to be shared across layers, but its binding mechanism restricts applicability to weights of varying shapes. ShareLoRA~\cite{song2024sharelora} introduces an asymmetric sharing mechanism where the matrix $A$ is shared across all layers, while the matrix $B$ is not. Although this approach significantly reduces the number of parameters by reusing $A$ across layers, it is relatively simplistic and lacks a detailed analysis of whether $B$ could also benefit from sharing.

To this end, we investigate the effects of partially sharing matrix $B$ while maintaining full sharing of matrix $A$ across all layers. We conduct preliminary experiments and show the results in Figure~\ref{fig:pre}. We observe that different sharing strategies yield different results, and in some cases, a smaller parameter size can lead to better performance. This suggests potential redundancies in $B$, pointing to a fine-grained sharing approach that reduces parameters while enhancing performance. Inspired by this, we propose a fine-tuning approach called \textbf{A}daptive \textbf{S}haring \textbf{Lo}w-\textbf{R}ank \textbf{A}daptation Across Layers (ASLoRA). We divide the training process into three stages: shared training, adaptive merging, and final optimization. In the shared training stage, the matrix $A$ is shared across all layers to capture global information while reducing the number of trainable parameters by half. Meanwhile, matrix $B$ remains unshared to capture the unique information of each layer. In the adaptive merging stage, to eliminate redundancy among the $B$ matrices of different layers and further reduce parameters, we merge these matrices based on their similarity. In the final optimization stage, the merged model structure is retained and further trained to ensure convergence and optimal performance. Compared to LoRA, ASLoRA combines global and local sharing, using fewer parameters and effectively alleviating the overfitting problem.
We conduct comprehensive experiments on multiple tasks and models, using RoBERTa-base for natural language understanding (NLU) tasks and LLaMA-2-7B for instruction tuning tasks. The experimental results show that ASLoRA achieves better performance with fewer parameters than LoRA, outperforming the baseline models across all instruction-following datasets. In summary, our contributions are as follows: 
\begin{itemize}[leftmargin=*, nosep=*]
    \item We experiment with different ways of sharing matrix $B$ while maintaining full sharing of matrix $A$ across all layers. We find that some strategies with fewer parameters perform better.
    \item We propose a parameter-sharing approach, ASLoRA, which combines global sharing with partial adaptive sharing to further enhance parameter efficiency.
    \item We compare ASLoRA with existing methods across multiple tasks, showing that it achieves higher parameter efficiency and superior performance.
\end{itemize}

\section{Related Work}
\subsection{Parameter-Efficient Fine-Tuning}
As transformer models scale up and downstream tasks increase, full fine-tuning poses significant computational challenges. To address this, parameter-efficient fine-tuning methods have emerged, which update only a small portion of the model’s parameters to achieve performance comparable to full fine-tuning. Prompt Tuning~\cite{DBLP:conf/emnlp/ShinRLWS20, DBLP:conf/www/ChenZXDYTHSC22} introduces task-specific prompts to adjust the model precisely, Adapter Tuning~\cite{DBLP:conf/icml/HoulsbyGJMLGAG19} adds lightweight adapters between model layers to drastically reduce resource consumption, and Prefix-Tuning~\cite{li-liang-2021-prefix} prepends a continuous, task-specific vector sequence to the model’s input. While these methods have shown remarkable effectiveness, fine-tuning large models still demands substantial computational resources, especially in resource-constrained environments.

\begin{figure*}[t!]
    \centering
    \includegraphics[width=\textwidth]{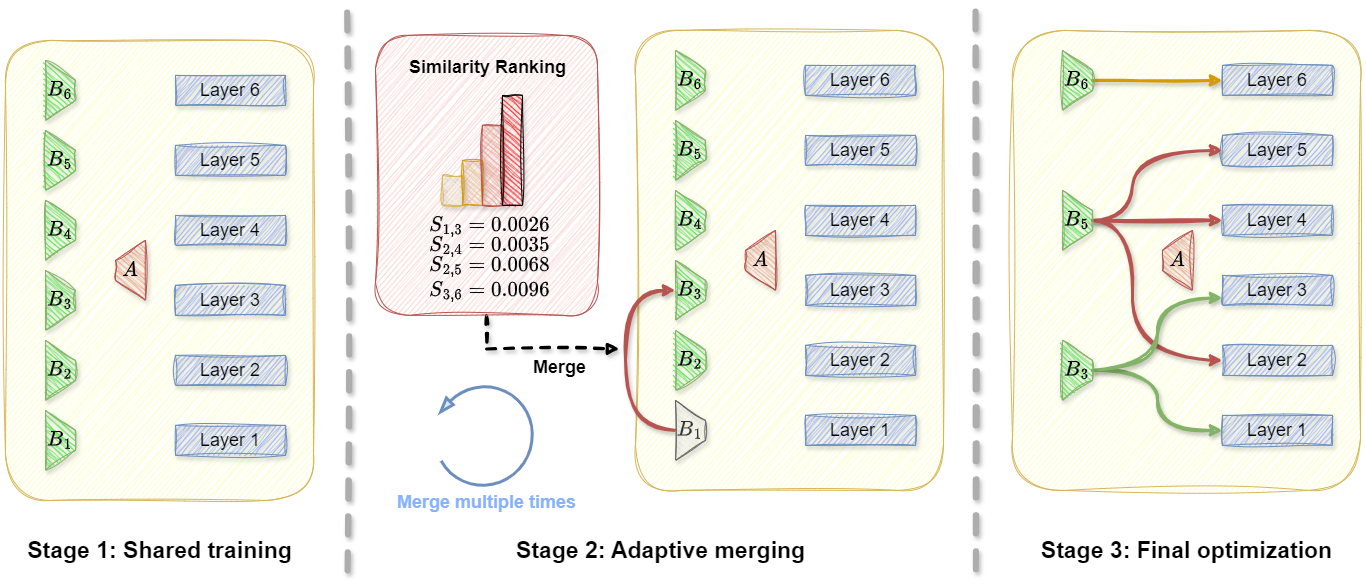}
    \caption{Illustration of ASLoRA, we present a six-layer model. First, all layers share matrix $A$ and enter the shared training phase on the left. The center shows the adaptive merging process, where the most similar $B$ matrices are merged each time based on their pairwise similarity. After several merges, the model moves to the final optimization phase on the right, with partial sharing of $B$ completed.}
    \label{fig:sample-image}
\end{figure*}

\subsection{Low-Rank Adaptation}
Low-Rank Adaptation (LoRA)~\cite{hu2022lora} using the product of two low-rank matrices to approximate weight updates. It is widely adopted due to its simplicity and lack of inference delay. Current improvements to LoRA focus primarily on enhancing performance and reducing parameter count. AdaLoRA~\cite{zhang2023adaptive} and IncreLoRA~\cite{DBLP:journals/corr/abs-2308-12043} improve LoRA by introducing higher ranks for more critical modules, but varying ranks across layers complicate multi-LoRA deployment. VeRA~\cite{DBLP:conf/iclr/KopiczkoBA24} reduces parameter count by sharing a frozen random matrix across layers and training two low-parameter vectors, but it affected performance. MELoRA ~\cite{ren-etal-2024-melora} connects multiple mini LoRAs to reduce parameter count while maintaining rank, at the cost of increased time complexity. PRoLoRA~\cite{wang-etal-2024-prolora} introduces shared and rotated enhancements within LoRA, effectively reducing parameters but remaining limited to internal LoRA interactions, thus unable to capture inter-layer dependencies. In contrast, our method employs a cross-layer parameter-sharing mechanism, effectively mitigating these limitations.
\subsection{Parameter Sharing}
Parameter sharing is widely used to reduce model memory requirements. Universal Transformer~\cite{dehghani2018universal} reduces parameter count by sharing all layers. ~\citeauthor{DBLP:conf/sustainlp/TakaseK23} introduced three cross-layer parameter sharing strategies that lower both parameters and computation demands. Subformer~\cite{reid-etal-2021-subformer-exploring} achieves significant parameter reduction without performance loss through middle-layer sharing and embedding factorization. LightFormer~\cite{lv-etal-2023-lightformer} uses SVD-based weight transfer and low-rank factorization for model compression and acceleration, while Relaxed Recursive Transformers~\cite{DBLP:journals/corr/abs-2410-20672} improves inference speed through cross-layer sharing via a recursive structure. Differently, our method focuses on PEFT scenarios, aiming to improve the parameter efficiency of LoRA models rather than directly optimizing transformer models.

\section{Method}
In this section, we will introduce ASLoRA, a adaptive method for sharing parameters across layers. In simple terms, we let $A$ share across all layers, and let $B$ share adaptively during training, reducing parameters while learning the information associated with each layer. We show our structure in Figure~\ref{fig:sample-image}.

\subsection{Preliminaries on Low-Rank Adapter}
LoRA freezes the original weight matrix $W_0$ and decomposes the weight update $\Delta W$ into two low rank matrices $B$ and $A$. The forward propagation process is shown in equation~\ref{eq:1}:
\begin{equation}
  \label{eq:1}
h=W_0 x+\Delta Wx=W_0 x + BAx.
\end{equation}
Here, $W_0 \in \mathbb{R}^{d \times d}$ is the pretrained weight matrix, $h$ is the output vector, $x \in \mathbb{R}^{d}$ is the input vector and $\Delta W = BA$ is the increment matrix during fine-tuning, where $A \in \mathbb{R}^{d \times r}$ and $r \ll d$. During training, $A$ is initialized with a Gaussian distribution, and $B$ is initialized as a zero matrix to ensure the initial increment $BA = 0$.

\subsection{Shared Training}
To reduce parameters while capturing global information across layers and local details for each layer, we need to share either $A$ or $B$. Considering that LoRA randomly initializes $A$, it means that $A$ is different for each layer, while $B$ is initialized to 0, meaning that $B$ is the same across all layers. Therefore, sharing $A$ while not sharing $B$ ensures that each layer's $B$ has the same initialization value, which facilitates measuring the changes in $B$. So we share $A$ across all layers and use a separate $B_i$ for each layer. The forward propagation process is shown in equation~\ref{eq:2}:
\begin{equation}
  \label{eq:2}
h_i=W_i x + B_iAx,
\end{equation}
where $i$ is the layer index of the model, $h_i$ is the output of layer $i$, $B_i$ represents the $B$ of the $i$-th layer. This equation indicates that the weight variation $\Delta W_i$ for each layer is obtained by matrix $A$ using the corresponding $B_i$. Shared $A$ is consistent across all layers, reducing redundancy in training and memory requirements. At the same time, the independent $B_i$ of each layer makes specific adjustments to the output to achieve differentiated feature transformation.

\subsection{Adaptive Merging}
After completing the $T_s$ steps of shared training, the model has learned  knowledge of different layers through $B$. In order to reduce the redundancy of $B$ and further reduce the parameters, we perform an adaptive merge of different $B$. In simple terms, we calculate the pairwise similarity between the $B$ matrices and merge $B$ with the highest similarity every $m$ steps.\\

\noindent\textbf{Average Weights.}
If we directly use the $B$ of step $t$ for similarity calculation, we would only observe the value of $B$ at the current step and fail to measure the overall changes in $B$ during the training phase. Therefore, we introduce the average weight to measure similarity. Specifically, the weight at step $t$ is equal to the average weight of the previous $t$ steps:
\begin{equation}
  \label{eq:3}
\overline {B_i^t} = \frac{1}{t} \sum_{k=1}^{t} B_{i}^k.
\end{equation}
Here, $i$ is the model layer index, $\overline {B_i^t}$ is the average weight of the $B$ for the layer $i$, $B_{i}^k$ is the weight of step $k$ of $B$, $t$ is the current step. By using average weights, we can better capture the overall $B_i$ information from the previous step, reducing randomness.\\

\noindent\textbf{Similarity Calculation.}
The L2 norm can effectively measures the overall distance between vectors and penalizes larger differences more significantly, so we use it to measure the similarity between the two pairs of $B$ at each layer, specifically:
\begin{equation}
  \label{eq:4}
S_{i, j}^{t} = \left \| \overline {B_i^t} - \overline {B_j^t} \right \|_2 = \sqrt{\sum_{k=1}^{n} (b_{i, k}^t-b_{j,k}^t)^2},
\end{equation}
where $S_{i, j}^t$ is the similarity between layer $i$ and layer $j$ matrices $B$, $b_{i, k}^t$ represents each element of layer $i$ matrix $B$. By using the L2 norm, we can effectively measure pairwise similarities between $B$-matrices and rank these similarities. From the equation~\ref{eq:4}, it can be seen that a smaller $S_{i, j}^{t}$ indicates a higher similarity. Each time, the two $B$-matrices with the highest similarity are selected for merging.\\

\noindent\textbf{Weight Merging.} Considering that the upper layers of the model contain more complex information~\cite{zhang2023adaptive}, we make the lower layers use the $B$ of the upper layers when merging. This ensures that more useful information is preserved after merging.

\subsection{Final Optimization}
After completing the merging of $B$, the model enters the final optimization phase. $A$ remains shared across all layers, while $B$ has undergone partial merged sharing. As a result, some layers share the same $B$, denoted as $\tilde{B} (i)$, representing the $B$ used in the $i$-th layer. The forward propagation formula is as follows:
\begin{equation}
  \label{eq:5}
h_i=W_i x + \tilde{B} (i)Ax.
\end{equation}
After this stage of training, the model has successfully converged. We summarize the detailed algorithm in Algorithm~\ref{Algorithm}.

\begin{algorithm}
\caption{: ASLoRA. $T$ is the total steps, $T_{s}$ is the number of steps that start merging, $m$ is the interval between merges, $N$ is the number of merges.}
\textbf{Input:} $T_{s}$, $m$, $N$
\begin{algorithmic}[1]
\STATE Share $A$ across all layers as equation (\ref{eq:2})
\FOR{\( t = 1, \dots, T \)}
    \IF{\(N > 0\) }  
        \STATE Update $\overline {B_i}$ by equation (\ref{eq:3})
        \IF{\(t>T_s\) and \((t-T_s) \% m\) == 0 }
            \STATE Calculate all $S^t$ by equation (\ref{eq:4})
            \STATE Sort all $S^t$ and find the minimal $S_{i,j}^t$
            \STATE Merge $B_i$ and $B_j$, $N \leftarrow N-1$
        \ENDIF
    \ENDIF
\ENDFOR
\end{algorithmic}
\label{Algorithm}
\end{algorithm}

As shown in Algorithm~\ref{Algorithm}, we first share $A$ across all layers and train the model according to equation~\ref{eq:2}, allowing $B$ to learn the information of each layer during this phase. After completing $T_s$ steps of training, we calculate the pairwise similarity between adjacent layers every $m$ steps, and merge the two layers with the lowest similarity. This process is repeated until $N$ merges are completed. Subsequently, we continue to train based on equation~\ref{eq:5}.

\subsection{Advantage Analysis}
\noindent\textbf{Global sharing $A$ has high adaptability.}
Because LoRA initializes $A$ randomly and $B$ with zeros, this initialization can affect the similarity calculations. By sharing $A$ across all layers, we can effectively eliminate the interference caused by the initialization values. Specifically, all layers' $B$ start with the same value (zero), and each $B$ propagates through the same $A$. This approach removes the influence of $A$ and the initialization values on $B$, leading to more reasonable and consistent similarity calculations for $B$.\\

\noindent\textbf{Partially sharing $B$ has high flexibility.}
We share $A$ across all layers to capture the shared knowledge across the entire model. Meanwhile, $B$ is partially shared based on the unique characteristics of each layer. This approach allows ASLoRA to capture both global knowledge and more fine-grained, layer-specific knowledge, providing greater flexibility. Especially when the model has more layers, this adaptive sharing strategy allows for a more flexible distribution of parameters.\\

\noindent\textbf{ASLoRA has high parameter efficiency.}
We share $A$ across all layers and merge $B$ during training. This approach can reduce the parameter size by at least half, and as the number of merges increases, the parameter size continues to decrease. As the number of model layers increases, the amount of parameters that can be reduced also increases.
\begin{table*}
  \centering
  \resizebox{\textwidth}{!}{
  \begin{tabular}{p{2cm}|r|ccccccc}
    \toprule
    
    \textbf{Method} & \#Params & \textbf{SST-2} & \textbf{MRPC} & \textbf{CoLA} & \textbf{QNLI} & \textbf{RTE} & \textbf{STS-B} & \textbf{Avg.}\\
    \midrule
    FF              & 125M   & \underline{94.8} & \textbf{90.2} & \underline{63.6} & 92.8 & 78.7 & \underline{91.2} & \underline{85.2} \\
    \midrule
    Adpt\textsuperscript{D}  & 0.3M   & 94.2 & 88.5 & 60.8 &93.1 & 71.5 & 89.7 & 83.0 \\
    Adpt\textsuperscript{D}  & 0.9M   & 94.7 & 88.4 & 62.6 & 93.0 & 75.9 & 90.3 & 84.2 \\
    LoRA            & 0.3M   & \textbf{95.1} & 89.7 & 63.4 & \textbf{93.3} & 78.4 & \textbf{91.5} & \underline{85.2} \\
    AdaLoRA         & 0.3M   & 94.5 & 88.7 & 62.0 & 93.1 & \textbf{81.0} & 90.5 & 85.0 \\
    DyLoRA          & 0.3M   & 94.3 & 89.5 & 61.1 & 92.2 & 78.7 & 91.1 & 84.5 \\
    PiSSA  & 0.3M   & 94.7 & 89.2 & \textbf{63.8} & 92.5 & 75.5 & 90.8 & 84.4 \\
    \textbf{ASLoRA} & \textbf{0.073M} & \underline{94.8} & \underline{90.0} & 63.3 & \underline{93.2} & \underline{79.8} & 91.1 & \textbf{85.4} \\
    \bottomrule
  \end{tabular}
  }

  \caption{Performance of various fine-tuning methods with RoBERTa-base models on 6 datasets of the GLUE benchmark. We report the Matthew’s correlation coefficient for CoLA, Pearson correlation coefficient  for STS-B and accuracy for other tasks. We also report the number of trainable parameters (\#Params) for each method. The best results for each dataset are shown in \textbf{bold}, the second-best results are \underline{underline}. Higher is better for all metrics in 6 datasets.}
  \label{tab:GLUEMain}
\end{table*}

\begin{table*}[ht]
\centering
\begin{tabular}{l|r|ccccc}
\toprule
\textbf{Method} & \#Params & \textbf{MMLU}  & \textbf{BBH} & \textbf{DROP} & \textbf{HEval} & \textbf{Avg.} \\
\midrule
w/o FT & -       & 45.96  & 32.04 & 31.55 & 12.20 & 30.44 \\
FT     & 7B      & \textbf{47.30} & 32.72 & 29.12  & 12.80 & 30.49 \\
\midrule
LoRA   & 33.6M   & 45.64  & 32.40 & \underline{32.46} & 15.09 & \underline{31.40} \\
QLoRA  & 33.6M   & 45.40  & 32.81 & 28.97 & \underline{15.24} & 30.61 \\
AdaLoRA & 33.6M  & 45.96  & \underline{32.85} & 31.94 & 14.02 & 31.19 \\
\textbf{ASLoRA} & \textbf{8.9M}    & \underline{46.21}  & \textbf{32.96} & \textbf{32.47} & \textbf{17.68} & \textbf{32.33} \\
\bottomrule
\end{tabular}
\caption{Results on instruction tuning, we present exact match scores for MMLU, DROP, and BBH, pass@1 for HumanEval(HEval). We also report the average score. With higher values indicating better performance. The best results for each dataset are shown in \textbf{bold}, the second-best results are \underline{underline}.}
\label{tab:Llama}
\end{table*}

\section{Experiments}
In this section, we evaluate the performance of ASLoRA in natural language understanding (NLU) and instruction tuning~\cite{chia-etal-2024-instructeval}. For NLU, we use RoBERTa-base~\cite{DBLP:journals/corr/abs-1907-11692} to test on the GLUE~\cite{wang-etal-2018-glue} dataset. For instruction tuning, we use LLaMA-2-7B as the large language model (LLM) backbone, trained on the alpaca dataset, and evaluate multiple metrics. Finally, we explore the advantages of adaptive merging.\\

\noindent\textbf{Baselines.}
We compare ASLoRA with popular parameter-efficient fine-tuning (PEFT) methods. To ensure a fair and comprehensive comparison, we replicate the experimental setups used in previous works and use their reported results. The baseline methods involved are:
\begin{itemize}[leftmargin=*, nosep=*]
\item\textbf{Full Fine-Tuning (FF)} - The base model is initialized with pre-trained weights and biases, and all parameters undergo gradient updates.
\item\textbf{Adapter Tuning} - 
\textbf{Adapter}\textsuperscript{\textbf{H}}~\cite{DBLP:conf/icml/HoulsbyGJMLGAG19} inserts two layers of adapters between the self-attention and feed-forward network modules, followed by a residual connection. We also compare three variants: \textbf{Adapter}\textsuperscript{\textbf{L}}~\cite{lin-etal-2020-exploring}, which applies adapter layers only after the MLP module, \textbf{Adapter}\textsuperscript{\textbf{P}}~\cite{pfeiffer-etal-2021-adapterfusion}, which applies adapters after the feed-forward layer, and \textbf{Adapter}\textsuperscript{\textbf{D}}~\cite{ruckle-etal-2021-adapterdrop}, which improves parameter efficiency by removing inactive adapter layers.
\item\textbf{LoRA}~\cite{hu2022lora} - LoRA parameterizes the incremental weight updates using low-rank matrices, making it a state-of-the-art PEFT method.
\item\textbf{DyLoRA}~\cite{valipour-etal-2023-dylora} - This method trains dynamic, search-free LoRA models to select the optimal rank.
\item\textbf{AdaLoRA}~\cite{zhang2023adaptive} - Based on singular value decomposition (SVD) and importance scores, AdaLoRA adaptively allocates different ranks to different modules of the model.
\item\textbf{PiSSA}~\cite{meng2024pissa} - PiSSA retains LoRA's architecture but initializes the low-rank matrices $A$ and $B$ with the principal components of the original weight matrix $W$, while storing the remaining components in a residual matrix.
\end{itemize}

\subsection{Natural Language Understanding}
\textbf{Models and Datasets.}  We validate our approach on the GLUE benchmark~\cite{wang-etal-2018-glue}, which includes a variety of natural language understanding (NLU) tasks, such as single-sentence classification, similarity and synonymous sentence tasks, and natural language reasoning tasks. We select RoBERTa-base model~\cite{DBLP:journals/corr/abs-1907-11692} for evaluation.\\

\noindent\textbf{Implementation Details.}  
In all experiments, we fine-tune $W_Q$ and $W_V$, with all data and models downloaded from huggingface. For the GLUE benchmark, we use the LoRA~\cite{hu2022lora} configuration, fine-tuning the RoBERTa-base model across 6 datasets. We set the rank to 8, and fine-tune all $W_Q$ and $W_V$ weights as well as the classification heads. For ASLoRA, since the Base model has only 12 layers and supports a maximum of 11 merges, we set the merge count to 7. We provide the hyperparameters in Table~\ref{tab:hyperparamsGLUE} in Appendix.\\

\noindent\textbf{Results.}  
The results are summarized in Table~\ref{tab:GLUEMain}. We report the number of all parameters except the classification header. For ASLoRA, we report the number of parameters after completing the merge. In the case of 7 merges, ASLoRA only uses 24\% (0.073M/0.3M) of the parameters, significantly reducing the parameter size, while surpassing all benchmark methods in average score. Although it fails to reach the leading position on a single data set, it ranks second on four data sets (SST-2, MRPC, QNLI and RTE), demonstrating its ability to diversify data sets while reducing the number of parameters. It maintains the advantages of stable performance and excellent generalization ability. Therefore, ASLoRA can significantly reduce the number of parameters and reach or exceed the performance of traditional methods under the condition of limited resources, which fully proves its feasibility and potential.

\subsection{Instruction Tuning}
\textbf{Models and Datasets.}
In this section, we use LLaMA-2-7B as the backbone LLM and train it using the alpaca dataset~\cite{taori2023stanford}, randomly selecting 2,000 samples as the development set. The alpaca dataset consists of 51K instruction-following examples generated by GPT-3.5~\cite{wang-etal-2023-self-instruct} , covering a variety of tasks and question formats, and it is designed to help the language model learn how to better understand and respond to instructions. We follow INSTRUCTEVAL~\cite{chia-etal-2024-instructeval} for evaluation, employing the MMLU~\cite{hendrycks2020measuring}, BBH~\cite{srivastava2022beyond}, DROP~\cite{dua2019drop}, and HumanEval (HEval)~\cite{chen2021evaluating} datasets.\\

\noindent\textbf{Implementation Details.}  
For all methods, we set the rank $r$ to 64. For ASLoRA, the maximum number of merges is set to 16. In terms of task setting, the MMLU uses 5-shot direct prompting, the BBH and DROP (dev) use 3-shot direct prompting, and the HEval uses 0-shot direct prompting, which reflects the complexity of different tasks and their requirements for model inference ability. During the training process, we use the AdamW optimizer, and train models for 3 epochs. The learning rate was based on a linear scheduling strategy, with an initial value of $3 \times 10^{-4}$. The batch size is set to 128. The above configuration ensures the consistency of experimental conditions and helps to comprehensively evaluate the performance of the model in each task. We provide the hyperparameters in Table~\ref{tab:hyperparamsLlama} in Appendix.\\

\noindent\textbf{Results.}  
The results are shown in Table~\ref{tab:Llama}, we find that ASLoRA uses only 26\% of the parameters required by other efficient fine-tuning methods, outperforms all baseline approaches on the BBH, DROP, and HEval datasets. While slightly underperforming full fine-tuning on the MMLU dataset, ASLoRA outperforms LoRA and its variants. Furthermore, ASLoRA achieves the highest average performance among the evaluated methods. These findings demonstrate that ASLoRA's integration of global and partial sharing mechanisms efficiently captures shared features across layers and allocates knowledge flexibly based on task demands. Consequently, ASLoRA significantly enhances model adaptability to diverse task complexities while preserving parameter efficiency, underscoring its promise in efficient fine-tuning.

\subsection{Further Analyses}
\textbf{Advantage of Adaptive Sharing.}
To further investigate the advantages of adaptive sharing, we conduct a comparative experiment against fixed sharing methods. In the fixed sharing method, the $B$ matrix is shared across every 2, 3, and 6 consecutive layers, whereas adaptive sharing merges 6, 8, and 10 times. These configurations are chosen for comparison because they maintain the same parameter counts, ensuring a fair evaluation. We conduct experiments on the MRPC, STS-B, SST-2, and QNLI datasets to evaluate the performance of adaptive sharing, with results presented in Table~\ref{fig:advantages}. The results indicate that adaptive sharing provides significant advantages across all configurations. For the 6-merging case (corresponding to sharing the $B$ matrix across every 2 layers), adaptive sharing yielded the largest performance improvement. Fewer merges enable adaptive sharing to allocate knowledge more flexibly, resulting in more diverse merging outcomes. However, when the number of merges increased to 10 (corresponding to sharing the $B$ matrix across every 6 layers), the performance advantage of adaptive sharing reduced. This is understandable, as increasing the number of merges limits the available options, reducing the diversity of adaptive sharing and making it closer to the fixed sharing method. In summary, adaptive sharing outperforms fixed sharing in terms of both parameter efficiency and performance, with its flexibility and adaptability offering significant advantages, particularly in configurations with fewer merges.\\
\begin{table}[ht]
\centering
\resizebox{1\columnwidth}{!}{ 
\begin{tabular}{l|r|ccccc}
\toprule
\textbf{Method} & \#Params & \textbf{MRPC}  & \textbf{STS-B} & \textbf{SST-2} & \textbf{QNLI} \\
\midrule
ASLoRA\textsuperscript{-adp} & 0.086M       & 88.48   & 90.80     &94.27      & 92.15 \\
ASLoRA     & 0.086M      & \textbf{90.20} & \textbf{90.92}      & \textbf{94.61}      & \textbf{92.93 }\\
\midrule
ASLoRA\textsuperscript{-adp}   & 0.061M   & 88.73  & \textbf{90.79}  &94.50     & 92.75 \\
ASLoRA  & 0.061M   & \textbf{88.97}  &90.73  & \textbf{94.84}      & \textbf{92.84}\\

\midrule
ASLoRA\textsuperscript{-adp}   & 0.037M   & \textbf{89.22}  & 90.42  & \textbf{94.27}      & 92.20 \\
ASLoRA  & 0.037M   & \textbf{89.22}  & \textbf{90.43}  & \textbf{94.27}      & \textbf{93.10}\\
\bottomrule
\end{tabular}
}
\caption{Performance on adaptive sharing and fixed sharing is compared. ASLoRA\textsuperscript{-adp} represents fixed sharing, with results corresponding to sharing matrix $B$ across every 2, 3, or 6 consecutive layers from top to bottom. These results are compared with adaptive sharing after merging 6, 8, and 10 times.}
\label{fig:advantages}
\end{table}

\begin{figure*}[t!]
    \centering
    \includegraphics[width=\textwidth]{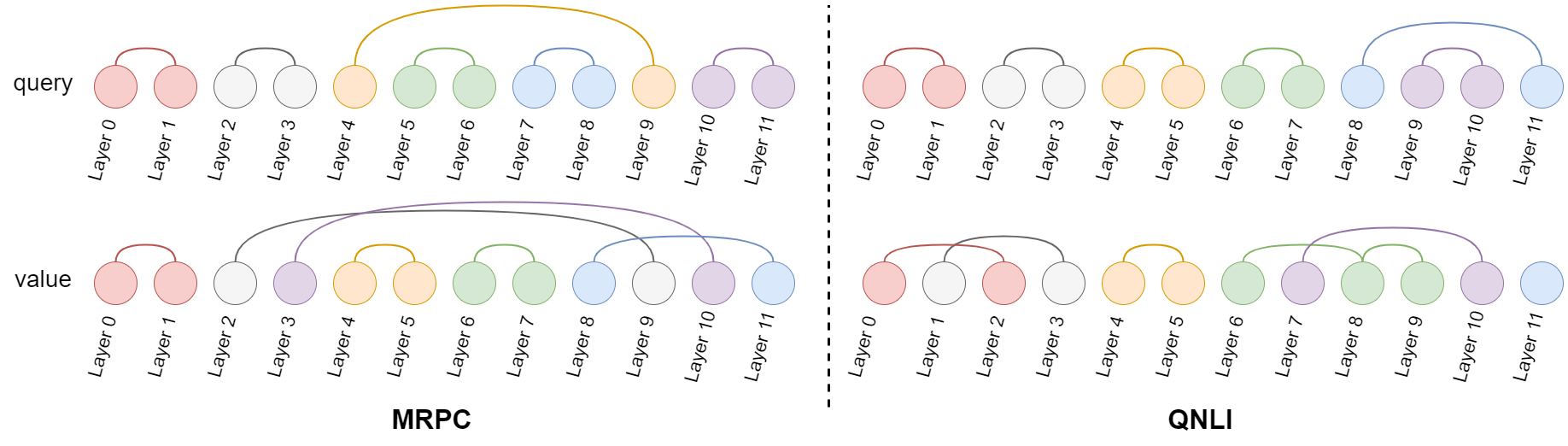}
    \caption{The allocation results of adaptive sharing on the GLUE Benchmark are presented. We set the merge times to 6 and report the sharing configurations of the query and value matrices. The same color represents sharing the same $B$ matrix. More results can be found in Figure~\ref{fig:distribute} in Appendix.}
    \label{fig:eye}
\end{figure*}

\begin{figure}[t!]
    \centering
    \begin{minipage}{0.25\textwidth}
        \centering
        \includegraphics[width=\textwidth]{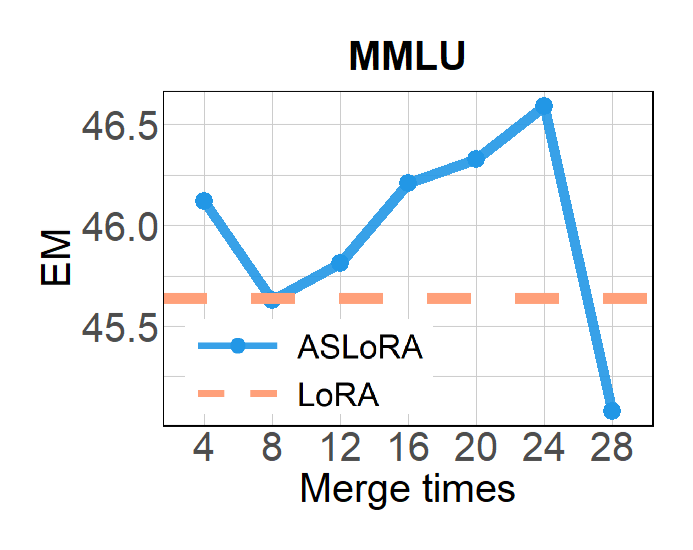}
    \end{minipage}%
    \begin{minipage}{0.25\textwidth}
        \centering
        \includegraphics[width=\textwidth]{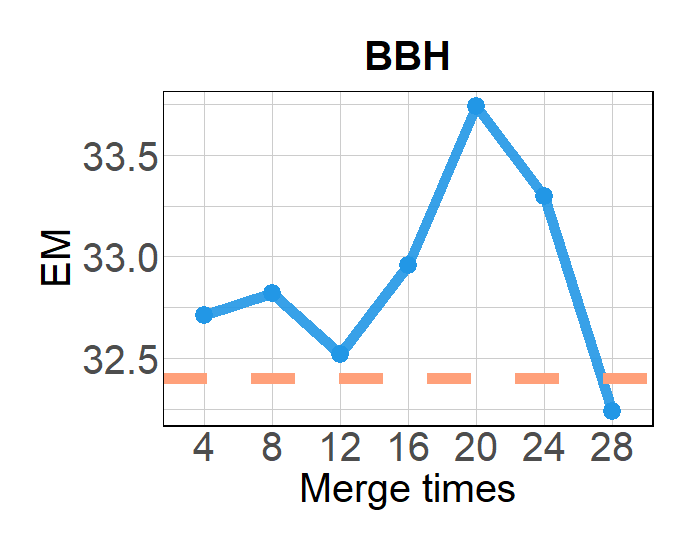}
    \end{minipage}

    \begin{minipage}{0.25\textwidth}
        \centering
        \includegraphics[width=\textwidth]{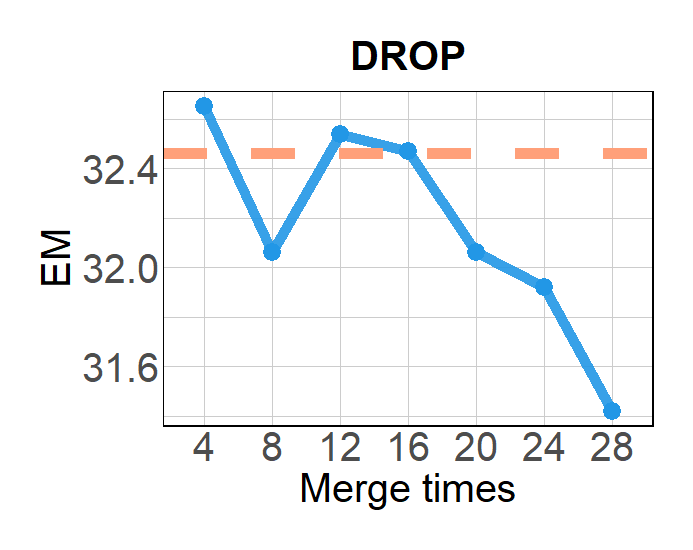}
    \end{minipage}%
    \begin{minipage}{0.25\textwidth}
        \centering
        \includegraphics[width=\textwidth]{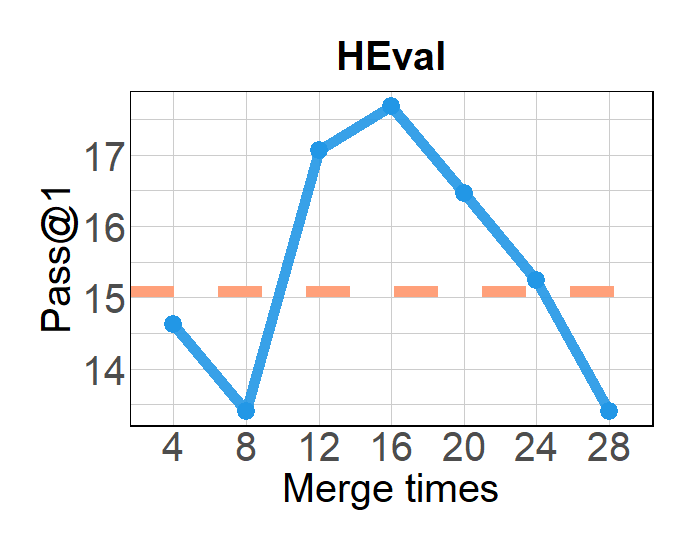}
    \end{minipage}%

    \caption{The effect of the number of merges on the results. Across these 4 datasets, for ASLoRA, we set the merge counts $N$ =\{4, 8, 12, 16, 20, 24, 28\} and conduct a comparative analysis with LoRA under the same rank $r$ setting.}
    \label{fig:mergetimes}
\end{figure}

\noindent\textbf{Shared Distribution.} 
To explore the impact of adaptive sharing on model structure, we conduct experiments on the RoBERTa-base model and reported the results of 6 merge iterations on the MRPC and QNLI datasets (as shown in Figure~\ref{fig:eye}). The results indicate that adaptive sharing achieves a more diversified allocation strategy. In the query matrix, the differences between adaptive sharing and fixed sharing are minimal, especially on the QNLI dataset, where the first 8 layers almost exclusively use adjacent two-layer sharing, with only slight differences appearing in the last four layers. This suggests that in the query matrix, inter-layer feature differences are small, and the performance of adaptive sharing and fixed sharing is similar. However, in the value matrix, the differences are more pronounced. Adaptive sharing exhibits a distinctly different sharing pattern, particularly on the QNLI dataset, where greater divergence is seen, especially in the sharing of layers 6, 8, and 9. Comparing MRPC and QNLI datasets, we find that adaptive sharing presents a more diverse allocation pattern on the QNLI dataset. This is because the QNLI dataset is larger and more complex than the MRPC dataset, providing a richer feature structure for the model to learn.
In summary, adaptive sharing can flexibly adjust the sharing strategy for each layer, significantly enhancing model performance and highlighting its advantages in model structure flexibility and task adaptability. In addition, ASLoRA can demonstrate more detailed allocation patterns on the value matrix and larger, more complex datasets.\\

\noindent\textbf{Impact of Merge Times.}
We explore the impact of different merge counts on the performance of instruction tuning, setting the merge counts $N = \{4,8,12,16,20,24,28\}$ and comparing the results with LoRA under the same rank settings, as shown in Figure~\ref{fig:mergetimes}. The results show that on the MMLU, BBH, and HEval datasets, performance improves initially with increasing merge counts but declines beyond a certain point. In most configurations, ASLoRA outperforms LoRA. Specifically, the best performance is achieved with $N=24$ for MMLU, $N=20$ for BBH, and $N=16$ for HEval. This indicates that the optimal merge count varies across datasets. The number of parameters decreases as the number of merges increases. Moderate merging helps mitigate overfitting, but excessive merging can harm performance. Conversely, fewer merges lead to an increase in the number of parameters, and too few merges may result in overfitting risks and lower parameter efficiency. On these four datasets, we also find that ASLoRA performs worse on the DROP dataset compared to the others. This may be due to the complexity of the tasks in this dataset, which makes it difficult for the reduced parameters to effectively capture its intricate features.

\section{Conclusion}
In this paper, we propose a parameter-efficient fine-tuning method called ASLoRA, which employs a cross-layer parameter-sharing mechanism combining global sharing and partial adaptive sharing strategies. This approach significantly enhances parameter efficiency during fine-tuning. Extensive experiments demonstrate that ASLoRA reduces the number of parameters while improving model performance across multiple datasets. 

\section{Limitations \& Future Work}
This work has the following limitations:
\begin{itemize}[leftmargin=*, nosep=*]
\item We introduce two hyper-parameters: the starting merge step and the interval between merges. Different configurations of these parameters may lead to performance variations. For the starting merge step, although we find that setting it to around an epoch yields good results, better patterns may exist. For the merge interval, we plan to introduce the global budget scheduler from AdaLoRA to design a more effective strategy for spacing between merges, thereby further optimizing performance.

\item The optimal number of merges varies across datasets. In future work, we plan to integrate a dynamic search algorithm to automatically determine the optimal number of merges, enhancing the model's adaptability and overall performance.

\item Our current approach is limited to inter-layer parameter sharing, which could potentially be complemented by incorporating intra-layer parameter sharing. Additionally, the method does not modify the internal structure of LoRA. In future work, our approach can be combined with other parameter-reduction methods that improve the LoRA structure (e.g., MELoRA) to achieve higher parameter efficiency.
\end{itemize}


\clearpage

\onecolumn
\appendix
\section{Hyper-parameters}
\label{sec:appendix}
The detailed hyper-parameter settings on the instruction tuning and GLUE datasets are listed in Table \ref{tab:hyperparamsLlama} and Table \ref{tab:hyperparamsGLUE}.
\begin{table}[h!]
\centering
\begin{tabular}{ll}
\toprule
\textbf{Hyper-Parameter} & \textbf{Value} \\ 
\midrule
Learning rate $\eta$     & 3e-4           \\ 
Batch size               & 128            \\ 
Number of epochs         & 3              \\ 
Max sequence length      & 256            \\ 
Rank $r$                 & 4              \\ 
Start Steps $T_s$         &400             \\
Merge interval $\mathcal W$ &10           \\
LoRA dropout             & 0.05           \\ 
LoRA alpha $\alpha$      & 16             \\ 
Trainable matrices       & $W_Q, W_V$     \\ 
LR scheduler             & Linear         \\ 
Warmup steps             & 100            \\ 
\bottomrule
\end{tabular}
\caption{The hyper-parameter settings for instruction tuning. We use the same settings as ~\cite{chia-etal-2024-instructeval}.}
\label{tab:hyperparamsLlama}
\end{table}

\begin{table}[h!]
\centering
\begin{tabular}{lcccccccc}
\toprule
\textbf{Hyper-Parameter} & \textbf{SST-2} & \textbf{MRPC} & \textbf{CoLA} & \textbf{QNLI} & \textbf{RTE} & \textbf{STS-B} \\ 
\midrule
Learning Rate $\eta$      & 5e-4           & 4e-4          & 4e-4          & 4e-4          & 4e-4          & 4e-4          \\ 
Batch Size                & 16            & 16           & 32            & 32           & 16          & 16          \\ 
Number of Epochs          & 60             & 30            & 80            & 25            & 25            & 40            \\ 
Weight Decay $\beta$      & 0.1            & 0.1           & 0.1           & 0.1           & 0.1           & 0.1           \\ 
Max Sequence Length       & 512            & 512           & 512           & 512           & 512           & 512           \\ 
Start Steps $T_s$             & 3000            & 320            & 400           & 1000           & 200           & 700      \\

Merge interval $\mathcal W$  &2000  &240  &500 &700  &100 &500 \\
Update Ratio $\lambda$    & 0.5            & 0.5           & 0.5           & 0.5           & 0.5           & 0.5           \\ 
Rank $r$                  & 8              & 8             & 8             & 8             & 8             & 8             \\ 
Alpha $\alpha$            & 16             & 16            & 16            & 16            & 16            & 16            \\ 
LR Scheduler              & Linear         & Linear        & Linear        & Linear        & Linear        & Linear        \\ 
Trainable Matrices        & $W_Q, W_V$     & $W_Q, W_V$    & $W_Q, W_V$    & $W_Q, W_V$    & $W_Q, W_V$    & $W_Q, W_V$    \\ 
Warmup Ratio              & 0.06           & 0.06          & 0.06          & 0.06          & 0.06          & 0.06          \\ 
Evaluation Metrics & Accuracy      & Accuracy      & Matthews      & Accuracy      & Accuracy      & Pearson       \\ 
\bottomrule
\end{tabular}
\caption{The hyper-parameter settings for GLUE.}
\label{tab:hyperparamsGLUE}
\end{table}

\section{Details of Datasets}
\subsection{GLUE Benchmark}
The GLUE~\cite{wang-etal-2018-glue} (General Language Understanding Evaluation) benchmark is a collection of natural language understanding tasks designed to evaluate the performance of language models in various practical applications. It provides a standardized platform for comparing how different models perform in understanding and processing human language. The GLUE benchmark includes nine tasks, each aiming to test different aspects of language understanding, such as text classification, sentence similarity, and reasoning. These tasks are MNLI~\cite{williams2018broad}(inference), SST-2~\cite{socher2013recursive} (sentiment analysis), MRPC~\cite{dolan2005automatically} (paraphrase detection), CoLA~\cite{warstadt2019neural} (linguistic acceptability), QNLI~\cite{rajpurkar2016squad} (inference), QQP (question-answering), RTE (inference), and STS-B~\cite{cer2017semeval} (textual similarity), we summarize their statistics in Table ~\ref{tab:gluedetail}.

\begin{table}[ht]
\centering
\resizebox{\textwidth}{!}{
\begin{tabular}{lccccccc}
\toprule
\textbf{Corpus} & \textbf{Task} & \textbf{\# Train} & \textbf{\# Val} & \textbf{\# Test} & \textbf{\# Labels} & \textbf{Metrics} & \textbf{Domain} \\
\midrule
\multicolumn{8}{c}{\textbf{Single-Sentence Tasks}}   \\
\midrule
CoLA & Acceptability & 8.55k & 1.04k & 1.06k & 2 & Matthews Corr. & misc. \\
SST-2 &Sentiment & 67.3k & 872 & 1.82k & 2 & Accuracy & Movie reviews \\
\midrule
\multicolumn{8}{c}{\textbf{Similarity and Paraphrase Tasks}}\\
\midrule
 MRPC & Paraphrase & 3.67k & 408 & 1.73k & 2 & Accuracy/F1 & News \\
 STS-B & Sentence similarity & 5.75k & 1.5k & 1.38k & 1 & Pearson/Spearman Corr. & misc. \\
 QQP & Paraphrase & 364k & 40.4k & 391k & 2 & Accuracy/F1 & Social QA \\
\midrule
\multicolumn{8}{c}{\textbf{Inference Tasks}}\\
\midrule
 MNLI & NLI & 393k & 19.65k & 19.65k & 3 & Accuracy & misc. \\
QNLI  & QA/NLI& 105k & 5.46k & 5.46k & 2 & Accuracy & Wikipedia \\
RTE & NLI & 2.49k & 277 & 3k & 2 & Accuracy & News \& Wikipedia \\
\bottomrule
\end{tabular}
}
\caption{Summary of GLUE benchmark tasks.}
\label{tab:gluedetail}
\end{table}

\subsection{Instruction Tuning}

\begin{itemize}[leftmargin=*, nosep=*]
    \item \textbf{MMLU}~\cite{hendrycks2020measuring}  evaluates models' knowledge and problem-solving skills across various fields. It tests performance in zero-shot and few-shot settings, making it highly challenging and closely aligned with human evaluation standards. The dataset covers 57 subjects, including STEM, humanities, and social sciences, with difficulty levels ranging from elementary to advanced professional. Each sample provides four choice of answers, and the task is to select the correct one.
    \item \textbf{BBH}~\cite{srivastava2022beyond} is a high-difficulty subset of the BIG-Bench benchmark, comprising 23 tasks designed to test scenarios that are challenging for current language models. These tasks include complex instructions such as navigation, logical reasoning, and fallacy detection.
    \item \textbf{DROP}~\cite{dua2019drop} is a math-focused reading comprehension benchmark that requires logical reasoning over Wikipedia-based passages. Models need to resolve references in the questions and perform discrete operations such as addition, counting, and sorting.
    \item \textbf{HumanEval}~\cite{chen2021evaluating} is a benchmark for evaluating code generation models. It includes 164 original programming tasks that assess language understanding, algorithms, and basic mathematical reasoning. Some problems resemble those found in basic coding interviews.
\end{itemize}

\newpage
\section{Analysis of Shared Distribution.}
\begin{figure}[h!] 
    \centering
    \includegraphics[width=\textwidth]{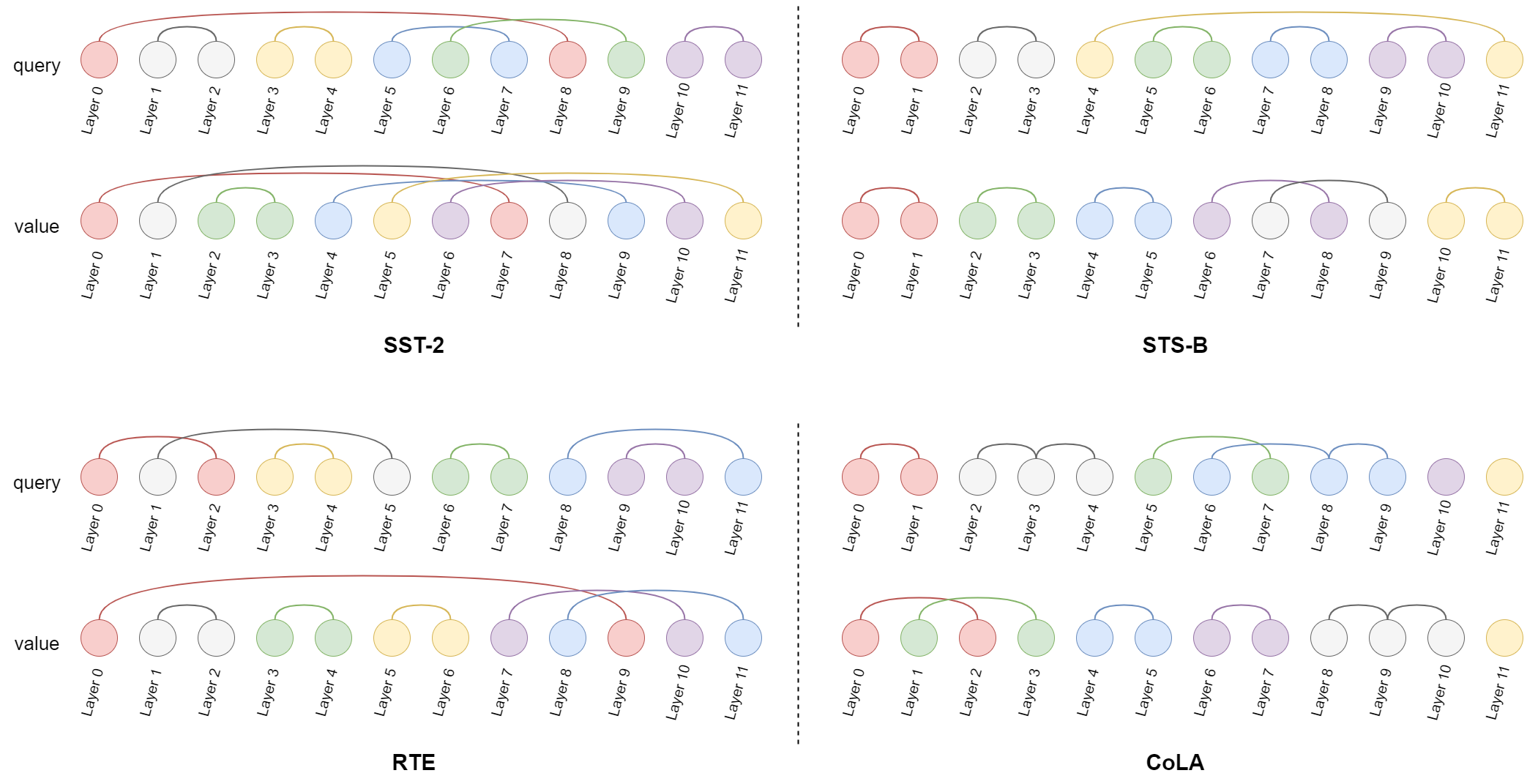}
    \caption{The allocation results of adaptive sharing on the GLUE Benchmark are presented. We set the merge times to 6 and report the sharing configurations of the query and value matrices. The same color represents sharing the same $B$ matrix.}
    \label{fig:distribute}
\end{figure}

In Figure~\ref{fig:distribute}, we present the results of additional adaptive allocations. For all datasets, the value matrix exhibits more diverse assignment patterns, with the complexity of these assignments varying across different datasets. Among them, SST-2 shows the most detailed allocation, likely due to its larger data size and more complex task.

\end{document}